\documentclass{article}
\usepackage{ijcai26}

\usepackage{times}
\usepackage{soul}
\usepackage{url}
\usepackage[hidelinks]{hyperref}
\usepackage[utf8]{inputenc}
\usepackage[small]{caption}
\usepackage{graphicx}
\usepackage{amsmath}
\usepackage{amsthm}
\usepackage{booktabs}
\usepackage{algorithm}
\usepackage{algorithmic}
\usepackage[switch]{lineno}

\usepackage{svg}
\usepackage{amssymb}

\usepackage{multirow}
\usepackage{booktabs}
\usepackage{subcaption}
\usepackage{xspace}
\usepackage{enumitem}

\def\name{HieraMix\xspace}
\title{\name: A Hierarchical MLP-Mixer for Large-Scale Traffic Forecasting}

\author{
Yongyao Wang$^{1,2,3,\dagger}$
\and
Xie Yu$^{2,3,\dagger}$
\and
Jingyuan Wang$^{1,2,3}$
\and
Jiahao Ji$^{2,3}$
\And
Chao Li$^{2,3}$\thanks{Corresponding author.}\\
\affiliations
$^1$State Key Laboratory of Complex \& Critical Software Environment (SKLCCSE)\\
$^2$School of Computer Science and Engineering, Beihang University, Beijing, China\\
$^3$MOE Engineering Research Center of Advanced Computer Application Technology, Beihang University, China\\
}

\begin{document}

\maketitle

\begin{abstract}
Traffic forecasting task is significant to modern urban management. 
Recently, there is growing attention on large-scale forecasting, as it better reflects the complexity of real-world traffic networks.
However, existing models often exhibit quadratic computational complexity, making them impractical for large-scale real-world scenarios. In this paper, we propose a novel framework, Spatio-Temporal \textbf{Hiera}rchical \textbf{Mix}er (\name), which leverages an all-MLP architecture for efficient and effective large-scale traffic forecasting. \name employs a hierarchical spatiotemporal mixing block to extract multi-resolution features through bottom-up aggregation and top-down propagation. Furthermore, an adaptive region mixer generates transformation matrices based on regional semantics, enabling our model to dynamically capture evolving spatiotemporal patterns for different regions. Extensive experiments conducted on four large-scale real-world datasets demonstrate that the proposed method not only achieves state-of-the-art performance but also exhibits competitive computational efficiency. 
\end{abstract}


\section{Introduction}

Traffic forecasting is a fundamental task in intelligent transportation systems, serving as the base for applications like congestion mitigation and route navigation. Typically, the performance depends on how accurately it can capture spatiotemporal correlations within traffic data. Early models like ARIMA~\cite{kumar2015short} rely on stationary assumptions, leading to limited representation ability. More recently, deep learning models, consisting of GNN-based models and transformer-based models et.~al, have become the mainstream, as these models automatically capture more appropriate and dynamic spatiotemporal correlations from large amounts of data. 

\begin{figure}[t!]
	\centering
	\begin{subfigure}{0.4\linewidth}
		\centering
		\includegraphics[width=\linewidth]{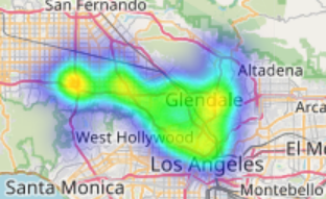}
            \caption{Attention of METR.}
		\label{subfig:i0}
	\end{subfigure}
	\centering
	\begin{subfigure}{0.4\linewidth}
		\centering
		\includegraphics[width=\linewidth]{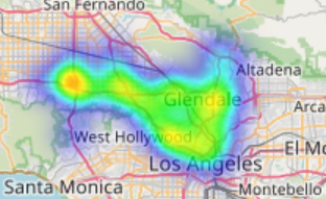}
		\caption{Similarity of METR.}
		\label{subfig:i1}
	\end{subfigure}
        \begin{subfigure}{0.4\linewidth}
		\centering
		\includegraphics[width=\linewidth]{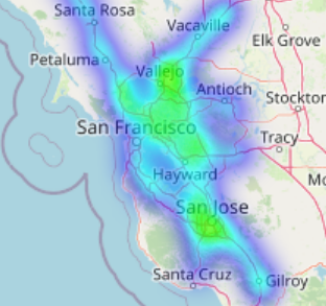}
		\caption{Attention of GBA.}
		\label{subfig:i2}
	\end{subfigure}
	\centering
	\begin{subfigure}{0.4\linewidth}
		\centering
		\includegraphics[width=\linewidth]{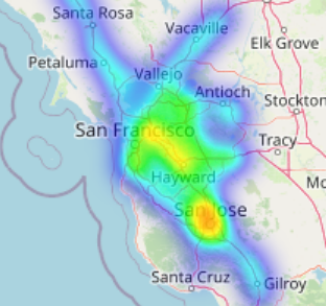}
		\caption{Similarity of GBA.}
		\label{subfig:i3}
	\end{subfigure}
    \caption{Attention performance on graphs of different scales. As the number of nodes increases in large-scale, the presence of significant noise dilutes the effectiveness.}
	\label{fig:intro}
\end{figure}

Despite the success of deep learning models, their compatible datasets are limited in scale compared to real-world traffic data. Applying these models to large-scale datasets remains a problem. Specifically, there are two key challenges: $\textbf{Scalability}$. Current deep-learning models exhibit quadratic computational complexity, causing an unacceptable increase in time and space costs as the data scale grows. 
\textbf{Effectiveness}. Existing models struggle to capture appropriate features when facing large-scale data. 
Each data sample is correlated with only a small subset of the dataset, while the remaining most act as noise. Current models often overlook this issue, indiscriminately constructing global correlations among samples, which scatters key features among substantial irrelevant noise.
As shown in Figure~\ref{fig:intro}, this issue is less evident in small datasets. However, as the scale of dataset increases, the proportion of noise grows significantly, making it increasingly difficult for models to capture key features effectively.

Pioneering works address above challenges from multiple perspectives. For scalability, there is a tendency to employ linear models, consisting of linear attentions~\cite{liang2023airformer} and MLP-Mixers~\cite{mlpst}. Among them, MLP-Mixers stand out for its lightweight and lower training costs.
However, above efforts process all samples indiscriminately, introducing massive noise through global feature integration, which limits their effectiveness on large-scale data. To mitigate this, some approaches~\cite{patchstg} group highly correlated samples into patches, enabling information exchange at the patch level. While this reduces computational complexity and partially addresses effectiveness, predefined patches fail to capture the evolving correlations of spatiotemporal data.
\textit{Thus, developing a linear model that dynamically identifies key features is essential.} Yet, explicitly defining key features is challenging, as they are shaped by complex and evolving spatiotemporal patterns.

\begin{figure}[t]
    \centering
    \includegraphics[width=\columnwidth, trim=0 1 0 5, clip]{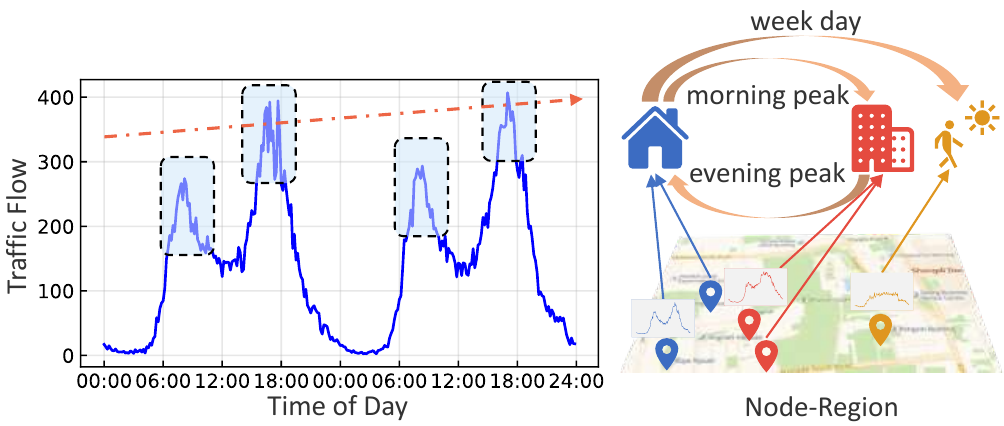}
    \caption{Spatiotemporal Hierarchy. At the macro-level, temporal data is driven by periodicity and trends (left), while spatial data is influenced by regional correlations, such as peak hours between residential and work areas (right). At the micro-level, neighboring samples exhibit similar values.}
    \label{fig:hierarchical}
\end{figure}

\label{intro_hierarchy}
Fortunately, spatiotemporal data exhibit a hierarchical structure, which becomes more evident with increasing data scale.
As illustrated in Figure~\ref{fig:hierarchical}, at the micro-level, local proximity prevails, with neighboring samples sharing similar values. At the macro-level, temporal data is driven by tendency and periodicity, while spatial data is influenced by semantic relevance.
In summary, spatiotemporal data at different data resolutions exhibit distinct dominant patterns, generating specific key features.
Such hierarchy is essential to address large-scale data problems, as these dominant patterns enable models to prioritize key samples, effectively filtering out irrelevant noise.
Moreover, extracting key features at each individual data resolution is more effective and manageable than directly capturing global key features from the entire dataset.
Motivated by the above analysis, we designed a linear model based on a hierarchical paradigm.

Technically, we proposed \name, a Hierarchical Spatio-Temporal Mixer model built entirely on Multi-Layer Perceptrons (MLPs). The core of \name is the spatiotemporal mixing block, which hierarchically extracts and integrates spatiotemporal features. This block comprises a temporal aggregation mixer for temporal local features extraction and a spatial cascade mixer modeling hierarchical region-level spatial features. Furthermore, we design an adaptive region mixer that dynamically mixes features based on regional semantics. As a linear model, \name is computationally efficient to large-scale datasets. By focusing on region-level dependencies, \name reduces the noise introduced by global modeling. Our contributions are summarized as follows:
\begin{itemize}
    \item We propose \name for large-scale traffic prediction. The model generates hierarchical spatiotemporal features through bottom-up aggregation and integrates these features with a top-down propagation mechanism.
    \item We design an adaptive region mixer that generates transformation matrices based on regional semantics, enabling to dynamically model spatiotemporal patterns for different regions.
    \item Extensive experiments are conducted on four large-scale datasets, and \name consistently outperforms state-of-the-arts, with average improvements of 5.32\%, 3.64\% and 3.55\% in MAE, RMSE, and MAPE, respectively.
\end{itemize}

\section{Related Work}

\paragraph{Traffic Forecasting} aims to predict traffic dynamics, e.g., traffic flow~\cite{wang2022traffic,stsl,ji2022stden}, traffic speed~\cite{bayesian,wang2016traffic}. Early deep learning-based methods leverage Recurrent Neural Networks~\cite{dcrnn,agcrn} or Temporal Convolutional Networks ~\cite{gwnet} to capture temporal dependencies, while Graph Neural Networks are used to model spatial dependencies~\cite{han2025bridging,ji2023multi}. Transformer-based models are widely applied to various urban spatio-temporal tasks, such as trajectory representation~\cite{start,BIGCity}, and general time-series forecasting~\cite{when,wang2020deep}. Recently, they also gain considerable popularity in traffic forecasting~\cite{pdformer,ji2025seeing,deng2025urban}. However, the intensive convolution operations of GNNs~\cite{l2gcn} and the quadratic complexity of transformers limit their ability on large-scale urban.

\paragraph{Large-Scale Traffic Forecasting} presents significant scalability challenges due to its extensive spatial scope, demanding models that are both efficient and effective. While some pioneering works reduce computational costs through graph decomposition~\cite{patchstg,STFT} or prior assumptions~\cite{GWT}, they may compromise information integrity. 
More recently, researchers begin to employ MLP-Mixers for modeling large-scale data, owing to their effectiveness and scalability.
MLPST~\cite{mlpst} first applies the MLP-Mixer architecture to grid-based data. TimeMixer~\cite{timemixer} performs node-level traffic forecasting by feature decomposition, but it lacks consideration of spatial correlations. TSMixer~\cite{tsmixer} and RPMixer~\cite{rpmixer} mix temporal and spatial dimensions, effectively capturing spatiotemporal dependencies. 
Despite their efficiency, these methods indiscriminately construct global correlations across all data samples. When facing large-scale data, it will overwhelm key features with substantial irrelevant noise. Recent works~\cite{wang2026tabsieve} also highlight the importance of reducing noisy context.

\begin{figure*}[t!]
	\centering
	\includegraphics[width=0.88\textwidth, trim=5 10 5 15, clip]{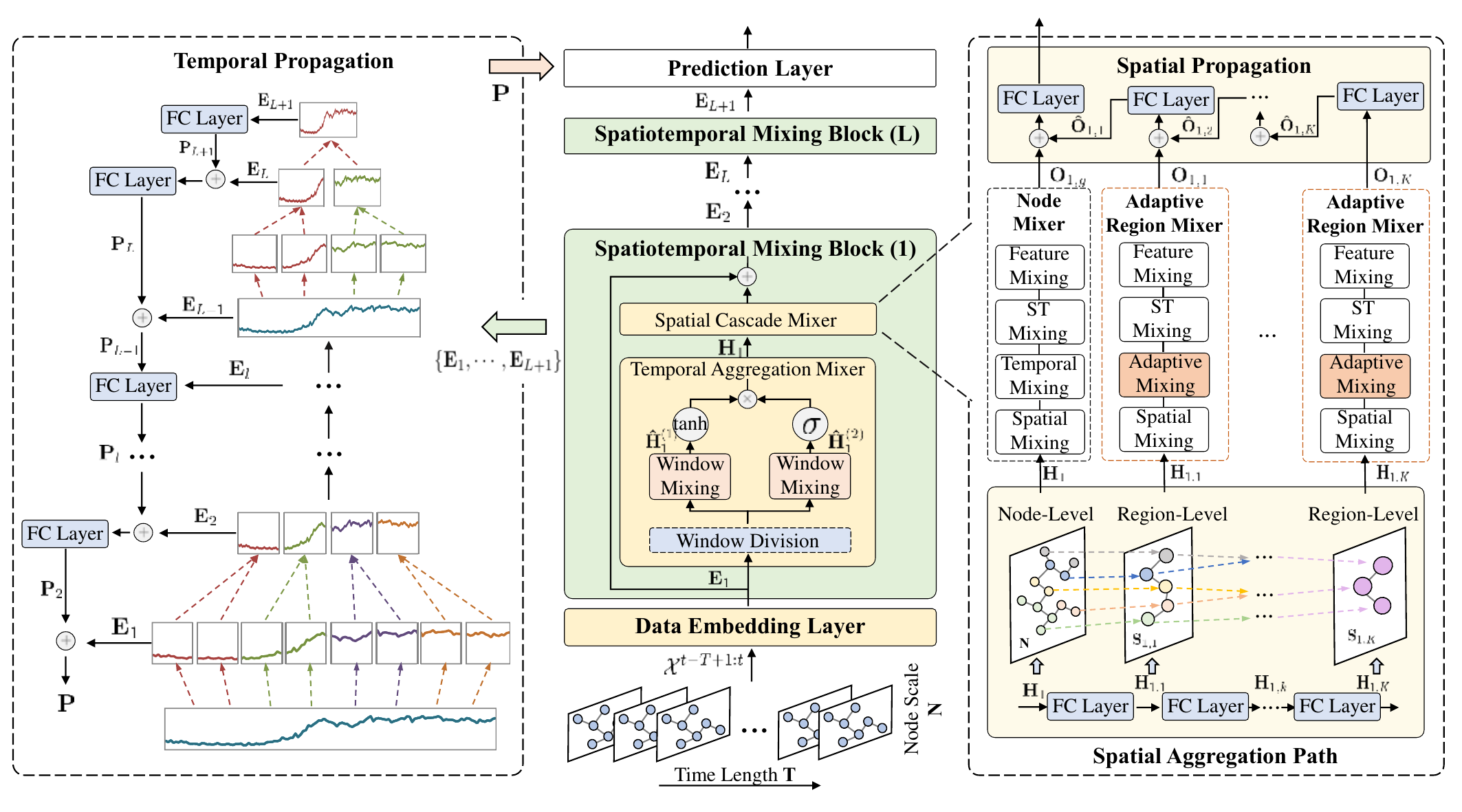}
	\caption{The overall framework of \name.}
	\label{fig:model}
\end{figure*}
\section{Preliminary}
\paragraph{Traffic Data.}
Traffic data consists of a series of numerical values collected and recorded by sensors in the urban traffic network. For the node $n$, $\mathbf{x}_n^h$ denotes traffic data at the $h$-th time slice. For a traffic network with $N$ nodes, $\mathbf{X}^h \in \mathbb{R}^{N}$ denotes overall traffic data at the $h$-th time slice. 
Further, $\mathcal{X}^{t-T+1\;:\;t}$ denotes a traffic data series from $t-T+1$ to $t$, where $\mathcal{X}^{t-T+1~:~t} = \left( \mathbf{X}^{t-T+1}, \cdots, \mathbf{X}^{t} \right) \in \mathbb{R}^{N \times T}$.

\paragraph{Traffic Forecasting.} 
Given a traffic data series $\mathcal{X}^{t-T+1~:~t}$ with its next $T^{\prime}$ steps $\mathcal{Y}^{t+1:t+T^{\prime}} = \mathcal{X}^{t+1:t+T^{\prime}}$. The objective of traffic forecasting is to learn a function $\mathcal{F}$ that predicts the future traffic data:
\begin{equation}
    \mathcal{\hat{Y}}^{t+1 \;: \; t+T^{\prime}} = \mathcal{F}(\mathcal{X}^{t-T+1 \; : \;t}, \Theta),
\end{equation}
where $\mathcal{\hat{Y}}^{t+1 \; : \;t+T^{\prime}}$ is the prediction to $\mathcal{Y}^{t+1 \; : \; t+T^{\prime}}$, and $\Theta$ denote the parameters to learn.

\section{Methodology}
Based on above analysis, the primary motivation for large-scale traffic forecasting is to leverage the hierarchy of spatiotemporal data.
Therefore, we propose our \name, a hierarchical MLP-Mixer model.
As shown in Figure~\ref{fig:model}, our \name consists of three parts.
Details of each part are described in the following sections.

\subsection{Data Embedding Layer}

In the data embedding layer, raw traffic data is first transformed into high-dimensional embeddings and then integrated with the spatial and temporal embeddings.

The spatial embeddings reflect the correlations between nodes in the traffic network, comprising static and dynamic components.
Specifically, the static component consists of node embeddings extracted from network topology through Node2Vec~\cite{node2vec}, denoted as $\mathbf{E}_{\text{static}} \in \mathbb{R}^{N \times d}$, where $d$ is the feature dimension. The dynamic component includes learnable vectors $\mathbf{E}_{\text{dynamic}} \in \mathbb{R}^{N \times d}$ to capture dynamic variations of each node. The overall spatial embeddings $\mathbf{E}_{sp} \in \mathbb{R}^{N \times d}$ are then obtained by:

\begin{equation}
\mathbf{E}_{sp} = \mathbf{E}_{\text{static}} + \mathbf{E}_{\text{dynamic}}.
\end{equation}

The temporal embeddings complement periodic patterns for traffic data, including two temporal components: minute-of-day~($1$ to $60 \times 24$) embeddings $\mathbf{E}_{\text{day}} \in \mathbb{R}^{T \times d}$ and day-of-week~($1$ to $7$) embeddings $\mathbf{E}_{\text{week}} \in \mathbb{R}^{T \times d}$. 
Subsequently, the temporal embeddings $\mathbf{E}_{te} \in \mathbb{R}^{T \times d}$ are computed as:
\begin{equation}
\mathbf{E}_{te} = \mathbf{E}_{\text{day}} + \mathbf{E}_{\text{week}}.
\end{equation}

Overall, the raw traffic data $\mathcal{X}^{t-T+1:t}$ is first transformed by a fully connected (FC) layer into $\mathbf{E}_{tr} \in \mathbb{R}^{N \times T \times d}$. Then, $\mathbf{E}_{tr}$ incorporates $\mathbf{E}_{sp}$ and $\mathbf{E}_{te}$ to generate the input embeddings $\mathbf{E}_1 \in \mathbb{R}^{N \times T \times d}$ for subsequent ST mixing blocks:
\begin{equation}
\mathbf{E}_1 = \mathbf{E}_{tr} + \mathbf{E}_{sp} + \mathbf{E}_{te}.
\end{equation}

\subsection{Spatiotemporal Mixing Block}
\paragraph{Motivation.}

Spatiotemporal data exhibits hierarchical characteristics, with dominated patterns vary with data resolutions. Our method aims to generate multi-resolution features through a bottom-up feature aggregation mechanism. The lower layers retain local characteristics, while the higher layers progressively incorporate global contextual features. This structured progression captures the transition between local and global properties and fully models patterns at different spatial scales. Specifically, we proposed the spatiotemporal mixing block (ST mixing block) based on the MLP-Mixer. By stacking multiple ST mixing blocks, hierarchical feature pyramids are generated efficiently.

\paragraph{Overall Structure.} 
As shown in Figure~\ref{fig:model}, \name consists of $L$ spatiotemporal mixing blocks, each ST mixing block comprises two sequential modules: a temporal aggregation mixer and a spatial cascade mixer. The temporal mixer captures local temporal features and aggregates them to generate macro-level features, forming a temporal feature pyramid across ST mixing blocks. The spatial cascade mixer focuses on spatial hierarchies, employing a cascade of mixers to construct a spatial feature pyramid and capture spatiotemporal patterns at specific time scale within each ST mixing block. Stacking multiple ST mixing blocks produces spatiotemporal features at various resolutions, which are integrated to obtain the final spatiotemporal representation.

Supposed $\mathbf{E}_l \in \mathbb{R}^{N \times T_{l-1} \times d}$ denotes the input of the $l$-th ST mixing block, where $N$ is the number of nodes and $T_{l-1}$ is the temporal length. Initially, the temporal aggregation mixer takes $\mathbf{E}_l$ as input and outputs $\mathbf{H}_l$. Then, $\mathbf{H}_l$ is processed by the spatial cascade mixer.

\subsubsection{Temporal Aggregation Mixer}

Firstly, $\mathbf{E}_l$ is divided into multiple $p$-length windows, where $\mathbf{E}_{l} \in \mathbb{R}^{N \times \left( T_{l} \times p\right ) \times d}$, and $T_{l} = \lceil T_{l-1} / p \rceil$ is the window size. Each window is processed by a shared structure which comprises two parallel window mixing MLPs. Each MLP applies two FC layers along the window-feature dimension~($p \times d$).

For each window mixing MLP, $\mathbf{E}_{l}$ is aggregated as $\mathbf{\tilde{E}}_{l} \in \mathbb{R}^{N \times T_{l} \times h}$ by the first FC layer, where $h$ is the intermediate feature dimension. Further, the positional embedding $\mathbf{E}_{\text{pe}} \in \mathbb{R}^{N \times T_{l} \times h}$ is added to $\mathbf{\tilde{E}}_{l}$ to identify the position information of each window. The second FC layer processes $\mathbf{\tilde{E}}_{l}$ and outputs $\mathbf{\hat{H}}_{l} \in \mathbb{R}^{N \times T_{l} \times d}$~:
\begin{equation}
    \mathbf{\hat{H}}_{l} = \text{FC}_2 \left( \phi \left( \text{FC}_1 \left(\mathbf{E}_{l} \right) + \mathbf{E}_{\text{pe}} \right ) \right),
\end{equation}
where $\phi(\cdot)$ is the activation function.
Considering there are two independent window mixing MLPs, let $\mathbf{\hat{H}}_{l}^{(1)}$ and $\mathbf{\hat{H}}_{l}^{(2)}$ denote the output of these two MLPs respectively.
To improve the expressive power of the model, we employ the gated mechanism~\cite{GTU} to generate the output $\mathbf{H}_{l} \in \mathbb{R}^{N \times T_{l} \times d}$ of the temporal aggregation mixer.
\begin{equation}
    \mathbf{H}_{l} = \text{tanh} \left (\mathbf{\hat{H}}_{l}^{(1)} \right) \odot \sigma \left( \mathbf{\hat{H}}_{l}^{(2)} \right ),
\end{equation}
where $\text{tanh}(\cdot)$ represents the tanh function, $\sigma(\cdot)$ denotes the sigmoid function, and $\odot$ is dot-product.

Unlike typical Mixers that preserve the input's shape, our temporal aggregation mixers integrate features within the same window.
When $p$ is small, values within each window exhibit gentle variations, dominated by the local similarity characteristics. As a result, feature aggregation can effectively capture local patterns and progressively adjust feature scale by reducing the number of windows across ST mixing blocks. For subsequent blocks, each window progressively contains richer information and reveals macro-level correlations like tendency.
After the $l$-th layer, the input $\mathbf{E}_{l} \in \mathbb{R}^{N \times \left(T_{l} \times p \right) \times d}$ is aggregated as $\mathbf{H}_{l} \in \mathbb{R}^{N \times T_{l} \times d}$.
\subsubsection{Spatial Cascade Mixer}\label{sec:SFB}

\emph{Overall Structure.} 
Similar to temporal correlations, spatial correlations are also multi-scale. At the node-level, correlations are dominated by local similarity. While at the region-level, they depend on the scale and semantics of regions.  
Therefore, we designed the spatial cascade mixer, consisting of a spatial aggregation path, a node mixer and $K$ adaptive region mixers corresponding to $K$ region scales. 

The spatial aggregation path consists of $K$ FC layers. Given the dynamic nature of spatial correlations and the lack of predefined regions, $\mathbf{H}_{l} \in \mathbb{R}^{N \times T_{l} \times d}$ is first progressively aggregated into a set of region-level features across different scales through the aggregation path, represented as $\left\{ \mathbf{H}_{l, 1}, \cdots, \mathbf{H}_{l, K} \right\}$. Suppose $\mathbf{H}_{l, k} \in \mathbb{R}^{S_k \times T_{l} \times d}$ is the input of the $k$-th adaptive region mixer, $S_k$ is the amount of regions in this mixer, and $\left(S_{k+1} < S_{k} < \cdots < S_1 < N\right)$.

Subsequently, each mixer processes the input along spatial, temporal, and feature dimensions. 
The specific structure of each mixer is illustrated in Figure~\ref{fig:model}.
Differed from the node mixer, the region mixer employs an adaptive mixing MLP to capture each region's unique features. As each region represents a distinct semantic unit, the adaptive mixing MLP ensures semantically similar regions share similar parameters, and vice versa. 
Each mixer generates its individual output, and the final output integrates features from both node-level and multiple region-level features.

\begin{figure}
    \centering
    \includegraphics[width=\columnwidth]{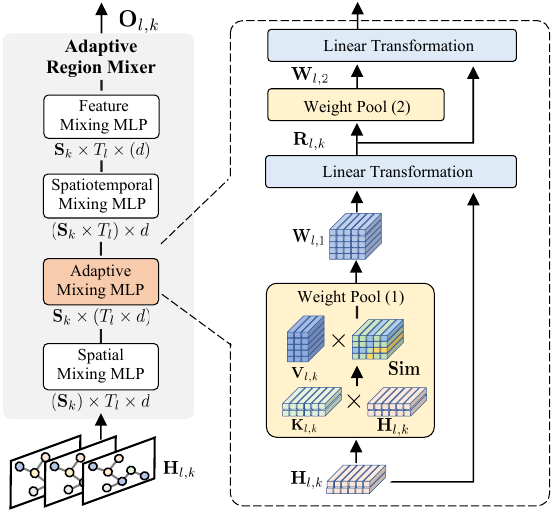}

    \caption{Structure of the Adaptive Region Mixer, where $\left( S_k \right) \times T_l \times d$ represents mixing along the spatial dimension $\left(S_k\right)$.}
    \label{fig:region}
\end{figure}
\emph{Adaptive Region Mixer.} 
As shown in Figure~\ref{fig:region}, the $k$-th adaptive region mixer processes $\mathbf{H}_{l, k} \in \mathbb{R}^{S_k \times T_{l} \times d}$ through four MLPs: the spatial mixing MLP, the adaptive mixing MLP, the spatiotemporal mixing MLP, and the feature mixing MLP to capture spatiotemporal dependencies at the corresponding scale. 
Notably, the adaptive mixing MLP is different as it employs a parameter pool to adaptively generate transformation matrices tailored to each region's semantics.

Specifically, the parameter pool contains $\mathbf{K}_{l, k} \in \mathbb{R}^{M_k \times T_{l}}$ as keys and $\mathbf{V}_{l, k} \in \mathbb{R}^{M_k \times T_{l} \times h}$ as base weights, where $h$ is the intermediate feature dimension and $M_k$ is the size of base weights. Then, the demanding parameters $\mathbf{W}_{l, k} \in \mathbb{R}^{S_k \times T_{l} \times h}$ are generated as follows:
\begin{equation}
\begin{aligned}
    & \mathbf{Sim} = \text{Softmax}(\mathbf{H}_{l, k} \ast \mathbf{K}_{l, k}^\top), \\
    & \mathbf{W}_{l, k} = \textstyle{\sum_{i=1}^{d}} \mathbf{Sim}^{(i)} \ast \mathbf{V}_{l, k},
\end{aligned}
\end{equation}
where $\ast$ denotes the matrix multiplication, $\mathbf{Sim} \in \mathbb{R}^{S_k \times d \times M_k}$ are weight scores, and $\mathbf{Sim}^{(i)}$ is the $i$-th tensor of $\mathbf{Sim}$ over the feature dimension $(d)$.

In the adaptive mixing MLP, both two FC layers are generated from the parameter pool, denoted as $\textbf{W}_{l,k,1}$ and $\textbf{W}_{l,k,2}$. Then, the input $\mathbf{H}_{l, k}$ is transformed as:
\begin{equation}
\begin{aligned}
    & \mathbf{R}_{l, k}^{(j)} = \mathbf{H}_{l, k}^{(j)\top} \ast \mathbf{W}_{l,k,1}^{(j)}, \\
    & \mathbf{H}_{l, k}^{(j)} := \mathbf{W}_{l,k,2}^{(j)} \ast \mathbf{R}_{l, k}^{(j)\top},
\end{aligned}
\end{equation}
where $\mathbf{H}_{l, k}^{(j)}$, $\mathbf{W}_{l,k,1}^{(j)}$ are both the $j$-th tensor over the spatial dimension, and $\mathbf{R}_{l, k} \in \mathbb{R}^{S_k \times d \times h}$ denotes the hidden state.

To increase the semantic distinctiveness of each region, we employ the parameter orthogonal loss (POL) during training. It minimizes the similarity between each $\textbf{W}_{l,k}^{(i)}$ and $\textbf{W}_{l,k}^{(j)}$ to decrease the semantic similarity among regions. Specifically, the POL in the $l$-th ST mixing block can be formalized as:
\begin{equation}
    \text{POL}_{l} = \textstyle{\sum_{k=1}^{K}} 
    \sum_{i=1}^{S_{k}} \sum_{j=1}^{S_{k}} \frac{1}{S_k^2}\text{CS}(\mathbf{W}_{l,k}^{(i)}, \mathbf{W}_{l,k}^{(j)}) ,
\end{equation}
where $\text{CS}(\cdot)$ represents the cosine similarity. The total orthogonal loss can be denoted as:
\begin{equation}
    \mathcal{L}_{\text{POL}} = \textstyle{\sum_{l=1}^L} \text{POL}_l.
\end{equation}

\textit{Output.} The spatial cascade mixer generates an output set, including $\left\{ \mathbf{O}_{l, g}, \mathbf{O}_{l, 1}, \cdots, \mathbf{O}_{l, K} \right\}$, where $\mathbf{O}_{l, g} \in \mathbb{R}^{N \times T_{l} \times d}$ belongs to the node mixer, and $\mathbf{O}_{l, k} \in \mathbb{R}^{S_k \times T_{l} \times d}$ belongs to the $k$-th adaptive region mixer.

\subsubsection{Hierarchical Feature Propagation}
Since \name processes spatiotemporal data hierarchically, it generates various feature pyramids: \textbf{(1)} Stacked ST mixing blocks form a feature pyramid across temporal resolutions, and the higher features reflect the trend of the lower features. \textbf{(2)} In each ST mixing block, the spatial cascade mixer generates a feature pyramid across spatial resolutions, and the higher regions reflect the interactions between lower nodes.
Therefore, the hierarchical feature propagation is proposed for feature integration. It contains two top-down paths: the spatial propagation and the temporal propagation.

\emph{Spatial Propagation.} The top-down spatial feature propagation starts with the output $\mathbf{O}_{l, K}$. It is transformed by an FC layer into $\hat{\mathbf{O}}_{l, K} \in \mathbb{R}^{S_{k-1} \times T_l \times d}$. 
As shown in Figure~\ref{fig:model}, $\mathbf{O}_{l, k}$ is transformed into $\hat{\mathbf{O}}_{l,k}$ as follows:
\begin{equation}
\begin{aligned}
    & \hat{\mathbf{O}}_{l,k} = \text{FC}\left( \mathbf{O}_{l, k} + \hat{\mathbf{O}}_{l, k+1} \right), \\
    & \mathbf{E}_{l+1} = \text{FC}\left( \mathbf{O}_{l, g} + \hat{\mathbf{O}}_{l, 1} \right) + \mathbf{E}_l,
\end{aligned}
\end{equation}
where $\textbf{E}_{l+1}$ denotes the final result of the spatial propagation, which serves as the input of the $(l+1)$-th ST mixing block.

\emph{Temporal Propagation.}
Stacked ST mixing blocks will generate a feature pyramid across temporal resolutions, denoted as $\left\{ \mathbf{E}_1, \cdots, \mathbf{E}_{l}, \cdots, \mathbf{E}_{L+1} \right\}$. Additionally, $\mathbf{E}_{L+1}$ is the output of the $L$-th ST mixing block. 
Similar as the spatial propagation, temporal propagation begins with $\mathbf{E}_{L+1}$, which is first projected by an FC layer into $\mathbf{P}_{L+1}$. 
As shown in Figure~\ref{fig:model}, $\mathbf{E}_l$ is transformed into $\mathbf{P}_l \in \mathbb{R}^{N \times T_l \times d}$ as follows: 
\begin{equation}
\begin{aligned}
    &\mathbf{P}_l = \text{FC} \left( \mathbf{E}_l + \mathbf{P}_{l+1} \right), \\
    &\mathbf{P} = \text{FC} \left( \mathbf{E}_1 + \mathbf{P}_{2} \right),
\end{aligned}
\end{equation}
where $\mathbf{P} \in \mathbb{R}^{N \times T \times d}$ denotes the final result.

\subsection{Prediction Layer}
The prediction layer, takes $\mathbf{P} \in \mathbb{R}^{N \times T \times d}$ and $\mathbf{E}_{L+1} \in \mathbb{R}^{N \times T_L \times d}$ as inputs and regresses the prediction values as:
\begin{equation}
    \mathcal{\hat{Y}}^{t+1\;:\;t+T^{\prime}} = \text{FC}_3 \left ( \phi \left ( \text{FC}_2 \left(\mathbf{P} + \text{FC}_1\left (\mathbf{E}_{L+1}\right ) \right ) \right) \right).
\end{equation}

To optimize the model, we employ the mean absolute error (MAE) as the regression loss, which is denoted as $\mathcal{L}_{REG}$.
The overall loss combines the regression loss and the parameter orthogonal loss, which can be formalized as $\mathcal{L}$:
\begin{equation}
    \mathcal{L} = \alpha \mathcal{L}_{REG} + \beta \mathcal{L}_{POL},
\end{equation}
where $\alpha$ and $\beta$ are hyper-parameters for weight balance.

\begin{table}[t!]
    \centering
        \begin{tabular}{c|ccc}
            \toprule
            \textbf{Datasets} & \textbf{Nodes} & \textbf{Time steps} & \textbf{Time Range} \\
            \midrule
            SD    & 716   & 35040 & 1/1/2019-1/1/2020 \\
            GBA   & 2352  & 35040 & 1/1/2019-1/1/2020 \\
            GLA   & 3834  & 35040 & 1/1/2019-1/1/2020 \\
            CA    & 8600  & 35040 & 1/1/2019-1/1/2020 \\
            \bottomrule
        \end{tabular}%
    \caption{Statistics of datasets.}
    \label{tab:datasets}%
  \end{table}%

\begin{table*}[t!]
  \centering
  \small
  \setlength{\tabcolsep}{4pt}
    \begin{tabular}{c|ccc|ccc|ccc|ccc}
    \toprule
    \multirow{2}{*}{\textbf{Model}} & \multicolumn{3}{c|}{\textbf{SD}} & \multicolumn{3}{c|}{\textbf{GBA}} & \multicolumn{3}{c|}{\textbf{GLA}} & \multicolumn{3}{c}{\textbf{CA}} \\
          & MAE   & RMSE  & MAPE  & MAE   & RMSE  & MAPE  & MAE   & RMSE  & MAPE  & MAE   & RMSE  & MAPE \\
    \midrule
    DCRNN~\cite{dcrnn} & 17.96 & 28.65 & 11.85 & 21.86 & 34.65 & 20.19 & 23.36 & 36.19 & 14.19 & 28.24 & 44.01 & 22.23 \\
    STGCN~\cite{stgcn} & 16.87 & 28.84 & 9.76  & 19.96 & 32.78 & 16.27 & 18.61 & 30.18 & 11.44 & 17.87 & 29.14 & 13.56 \\
    ASTGCN~\cite{astgcn} & 19.99 & 32.19 & 12.99 & 22.97 & 36.35 & 20.39 & 21.46 & 34.21 & 13.29 & \textendash     & \textendash     & \textendash \\
    GWNET~\cite{gwnet} & 16.93 & 27.97 & 9.74  & 19.73 & 31.99 & 17.54 & 18.94 & 30.16 & 11.58 & 19.73 & 31.32 & 15.42 \\
    STGODE~\cite{stgode} & 17.55 & 28.67 & 11.25 & 19.73 & 32.45 & 16.28 & 18.86 & 30.49 & 11.81 & 17.76 & 29.13 & 13.37 \\
    DSTAGNN~\cite{dstagnn} & 17.37 & 27.68 & 11.27 & 20.14 & 32.63 & 16.37 & 18.62 & 30.00    & 11.20  & \textendash     & \textendash     & \textendash \\
    D2STGNN~\cite{d2stgnn} & 16.49 & 26.59 & 9.62  & 18.56 & \underline{31.12} & \underline{13.05} & \textendash     & \textendash     & \textendash     & \textendash     & \textendash     & \textendash \\
    HIEST~\cite{hiest} & 17.94 & 29.03 & 10.33 & 22.50  & 35.75 & 14.02 & 22.37 & 35.26 & 11.84 & 22.06 & 35.03 & 14.18 \\
    PDFormer~\cite{pdformer} & 15.84 & 25.91 & 9.70   & \textendash     & \textendash     & \textendash     & \textendash     & \textendash     & \textendash     & \textendash     & \textendash     & \textendash \\
    DGCRN~\cite{dgcrn} & \underline{15.60}  & \underline{25.90}  & 9.93  & 19.11 & 31.21 & 14.17 & \textendash     & \textendash     & \textendash     & \textendash     & \textendash     & \textendash \\
    TESTAM~\cite{testam} & 18.51 & 30.42 & 10.42 & 19.57 & 32.32 & 13.94 & \textendash     & \textendash     & \textendash     & \textendash     & \textendash     & \textendash \\
    \midrule
    STID~\cite{stid}  & 18.21 & 29.97 & 10.18 & 20.82 & 34.43 & 13.19 & 19.83 & 32.33 & 10.38 & 19.11  & 31.45 & 11.68 \\
    FreTS~\cite{frets} & 23.95 & 37.12 & 14.89 & 26.87 & 40.55 & 18.39 & 27.76 & 42.00    & 15.85 & 27.58 & 42.50  & 18.47 \\
    TSMixer~\cite{tsmixer} & 16.81 & 29.79 & 10.52 & 19.93 & 34.83 & 16.56 & 17.86 & 30.89 & 10.96 & 16.78 & 29.42 & 12.43 \\
    TimeMixer~\cite{timemixer} & 17.15 & 28.30  & 9.74  & 21.74 & 35.15 & 13.85 & 19.76 & 31.96 & 10.81 & 18.16 & 30.11 & 11.61 \\
    RPMixer~\cite{rpmixer} & 16.72 & 29.64 & 10.63 & 19.05 & 32.75 & 15.26 & 17.57 & 30.42 & 10.56 & 16.75 & 29.28 & 12.36 \\
    PatchSTG~\cite{patchstg} & 15.93 & 28.19 & 9.69 & 18.94 & 33.02 & 13.17 & 17.53 & 29.72 & 10.02  & 16.69     & 28.58    & 11.50 \\
    EiFormer~\cite{eiformer} & 17.90  & 32.06  & 10.79  & 20.39  & 35.35  & 15.58  & 18.91  & 33.05  & 11.00  & 18.22  & 32.07  & 12.89  \\
    GSNet~\cite{gsnet} & 16.65  & 27.33  & 9.62  & 19.37  & 31.98  & 13.15  & 18.81  & 30.65  & 10.09  & 17.61  & 29.15  & 11.15  \\
    LSTNN~\cite{lstnn} & 15.83  & 27.08  & \underline{9.59}  & \underline{18.48}  & 31.59  & 13.11  & \underline{17.48}  & \underline{29.71}  & \underline{9.95}  & \underline{16.58}  & \underline{28.24}  & \underline{11.05}  \\
    \midrule
    \textbf{\name} & \textbf{14.80} & \textbf{25.06} & \textbf{9.22} & \textbf{17.73} & \textbf{30.67} & \textbf{12.71} & \textbf{16.45} & \textbf{28.03} & \textbf{9.53} & \textbf{15.55} & \textbf{27.05} & \textbf{10.55} \\
    \bottomrule
    \end{tabular}%
    \caption{Prediction performance of each model.The best results are highlighted in \textbf{bold}, while the second-best results are \underline{underlined}. Additionally, due to the limited scalability, some methods fail to handle large-scale datasets, with their results denoted as a dash (\textendash) in the table.}
  \label{tab:main}%
\end{table*}%

\section{Experiments}

\subsection{Experiment Setup}
\paragraph{Datasets.} 

We evaluate \name on four datasets within LargeST~\cite{liu2024largest}, including SD, GBA, GLA, and CA. 
Details of each dataset are described in Table~\ref{tab:datasets}.
Following previous works, we aggregate the time intervals from 5-minutes into 15-minutes. We divide each dataset according to four seasons, and then splite the training, validation and test sets with the ratio of 6:2:2 in each season.  

\paragraph{Baselines.}
The \name is compared with 20 baseline models, including 11 \textit{Previous SOTAs} and 9 \textit{Scalable Models for Large-Scale Settings}.

\paragraph{Experiment Details.}
All experiments are conducted on Ubuntu 20.04.6 LTS with an NVIDIA RTX A6000 48GB GPU, based on the LibCity~\cite{libcity}. Our \name takes 12 historical time slices as input to predict the next 12 time slices. We use the mean absolute error (MAE), the root mean square error (RMSE) and the mean absolute percentage error (MAPE) as evaluation metrics.

\subsection{Performance Comparisons}
Table.~\ref{tab:main} presents the overall performance of our model and all the baselines with respect to MAE, RMSE, and MAPE on four datasets. Additionally, we run all models five times and report the mean results. 
Overall, \name consistently outperforms other baselines on all four datasets. Specifically, \name obtains average improvements of 5.32\%, 3.64\% and 3.55\% beyond the best baselines in MAE, RMSE and MAPE, respectively. Furthermore, based on the experimental results, we can make the following observations:

\textit{Hierarchical Processing Benefits Performance.} 
In most cases, the performance of scalable baselines is much inferior to previous transformer-based models. However, \name enables a linear MLP-Mixer model to outperform complex transformer-based baselines across all four datasets. This is due to hierarchical processing, which allows the model to learn dominant features at different scales, generating more effective representations.

\begin{figure}[t]
	\centering
	\includegraphics[width=\columnwidth, trim=5 10 5 10, clip]{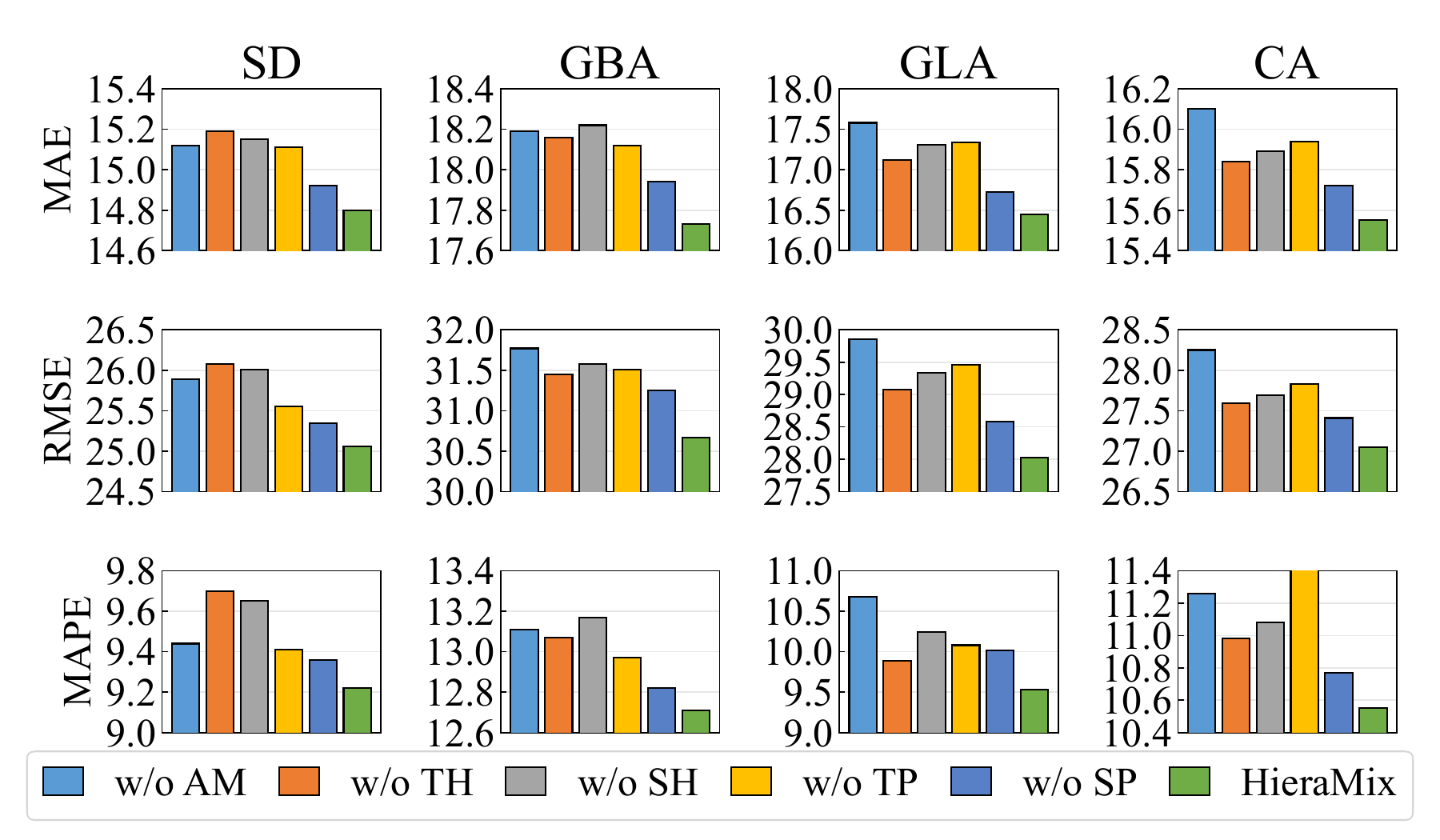}
	\caption{Results of ablation study.}
	\label{fig:ablation}
\end{figure}

\textit{Advantages of Scalability.} 
The experiments validate the scalability challenges inherent in previous SOTAs. Several GNN- and transformer-based models are unable to complete the experiments on larger datasets due to prohibitive computational and memory costs. In contrast, \name demonstrates excellent scalability, efficiently handling all datasets.

\textit{Advantages of Robustness.} 
The performance gap between \name and the second-best baseline widens with larger datasets. On the smallest dataset, 
\name outperforms the second-best baseline by 2.85\% in MAPE. On the largest dataset, this gap increases to 4.52\% in MAPE. This demonstrates \name's robustness with large-scale datasets.

\subsection{Ablation Study}
We conduct ablation studies to analyze the effectiveness of each component in \name. 
These five variants are listed as below:
\textbf{(1)} \textit{w/o AM:} Replaces the adaptive mixing MLP with a standard mixing MLP.
\textbf{(2)} \textit{w/o TH:} Removes the modeling of temporal hierarchy, i.e., the time length in each ST mixing block remains the same. 
\textbf{(3)} \textit{w/o SH:} Removes the modeling of spatial hierarchy, i.e., the number of nodes in the spatial cascade mixer remains the same. 
\textbf{(4)} \textit{w/o TP:} Removes the temporal propagation. 
\textbf{(5)} \textit{w/o SP:} Removes the spatial propagation.
Ablation results are shown in Figure~\ref{fig:ablation}, which can be concluded as: 

\textit{Benefits Brought by Hierarchical Modeling.} \textit{w/o SH} and \textit{w/o TH} are both inferior to \name, which demonstrates the effectiveness of hierarchical modeling in both spatial and temporal dimensions. Additionally, \name performs better than \textit{w/o AM}, because adaptive mixing improves semantic differences between regions, resulting in more structured hierarchical features.

\textit{Effectiveness of Feature Propagation.} As \name outperforms \textit{w/o TP} and \textit{w/o SP}, it indicates that features at different scales contain distinct semantic information. The feature propagation enables multi-scale features to complement each other, providing more comprehensive representations.

\begin{figure}[t]
	\centering
	\includegraphics[width=\columnwidth]{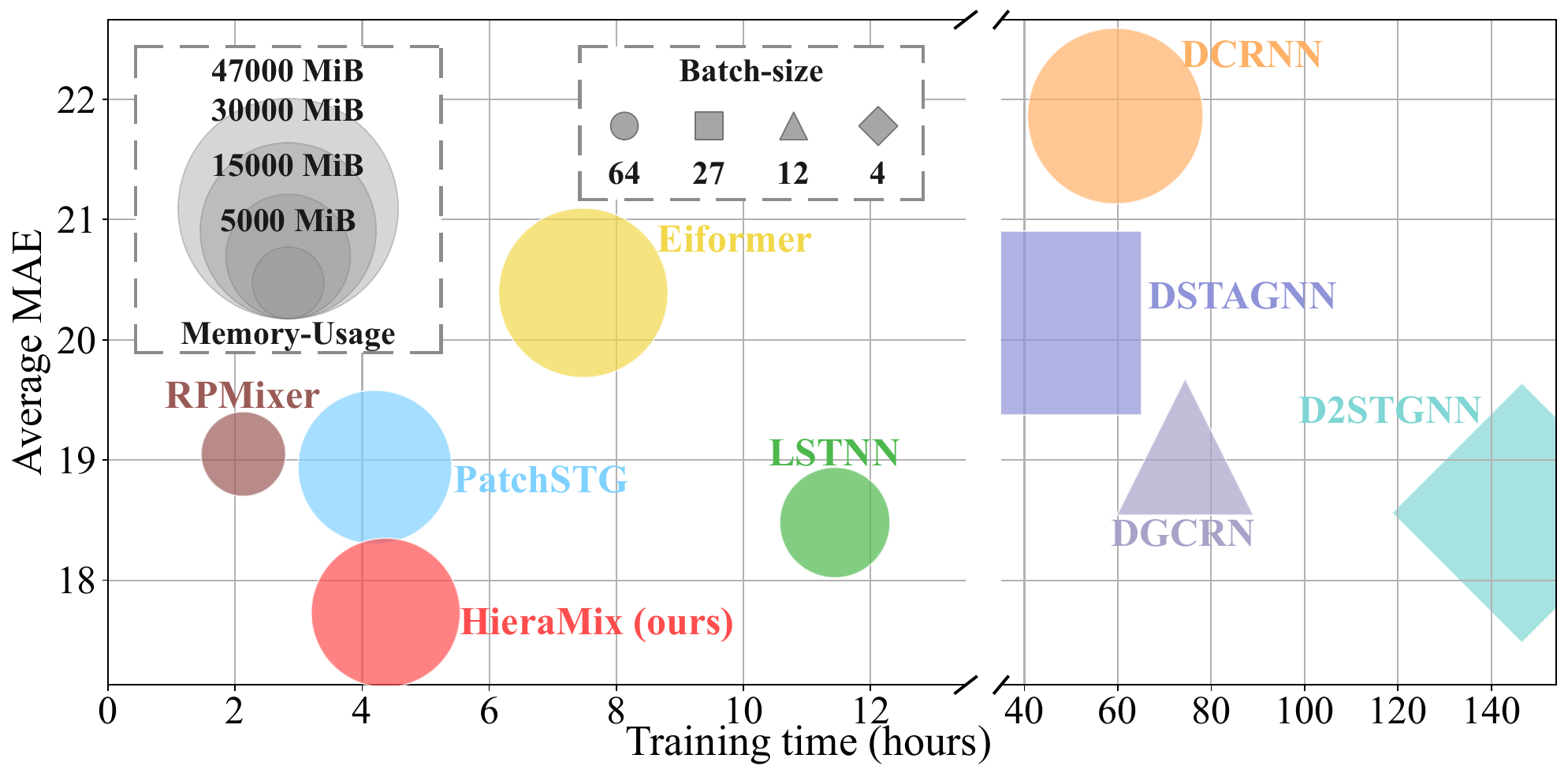}
	\caption{Performance and efficiency comparison.}
	\label{fig:efficiency}
\end{figure}
\subsection{Efficiency Analysis}
In this section, we evaluate \name and baselines in terms of performance and efficiency, and present results on MAE and the total training time on the GBA dataset. As shown in Figure~\ref{fig:efficiency}, \name significantly outperforms linear baselines and even surpasses SOTA transformer-based models (e.g., PatchSTG, Eiformer). Meanwhile, \name maintains efficiency comparable to mainstream scalable models (e.g., RPMixer, LSTNN). This demonstrates that \name balances performance and efficiency, offering excellent scalability on large-scale datasets. 

Further, we analyze the computational cost of \name. Specifically, the mixing operation mainly consists of two linear transformations with a time complexity of $O(dhNT)$. Generating adaptive transformation weights incurs a complexity of $O(MdhNT)$, and applying these weights for linear transformations adds another $O(dhNT)$. Consequently, the total complexity is $O(LKMdhNT)$, where $L$ and $K$ represent the number of ST mixing blocks and region scales, respectively. This linear complexity with respect to $N$ ensures scalability for large-scale spatiotemporal prediction.

\begin{figure}[t]
    \centering
        \includegraphics[width=\linewidth]{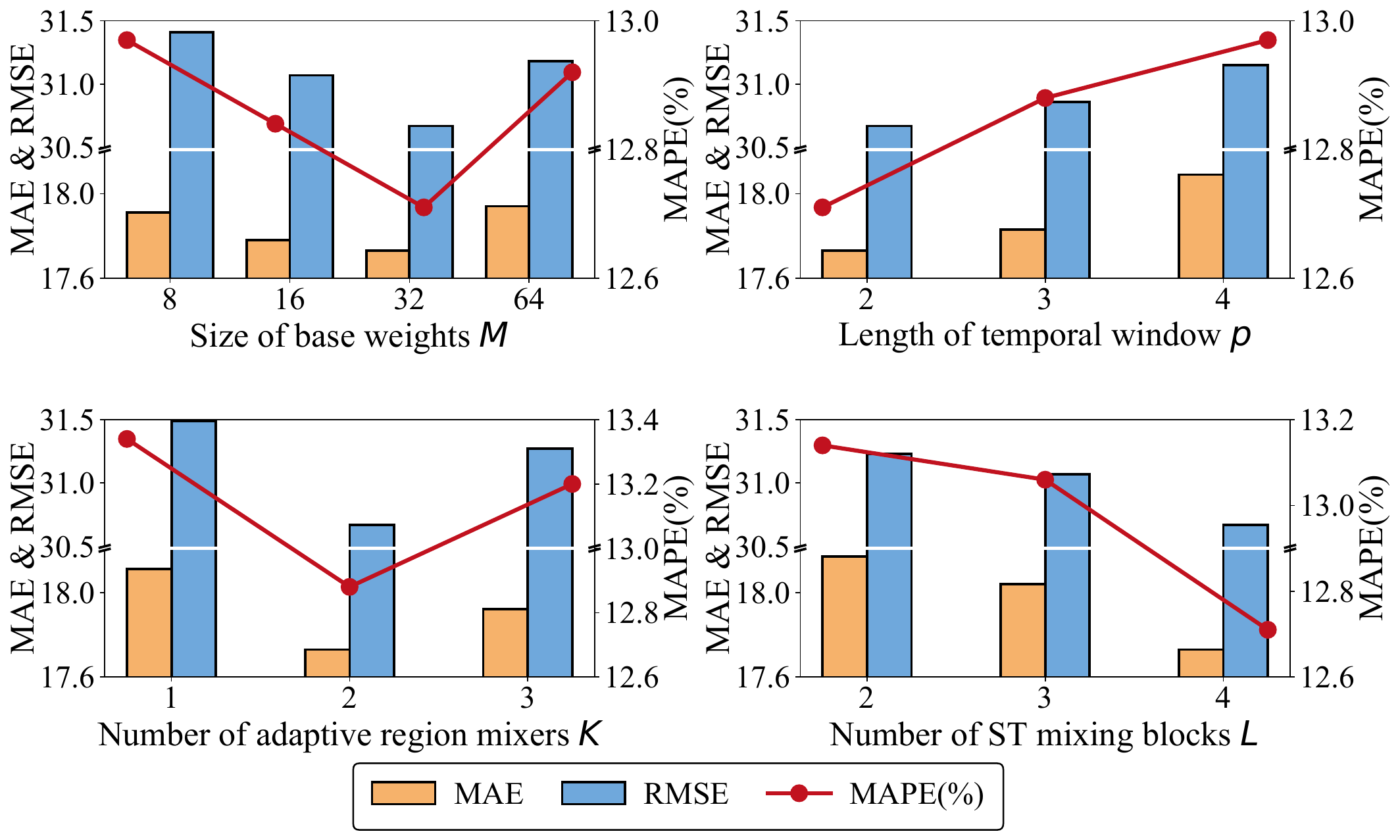}
    \caption{Sensitivity experiment results.}
    \label{fig:param}
\end{figure}

\begin{figure}[t]
	\centering
	\begin{subfigure}{0.49\linewidth}
		\centering
		\includegraphics[width=\linewidth]{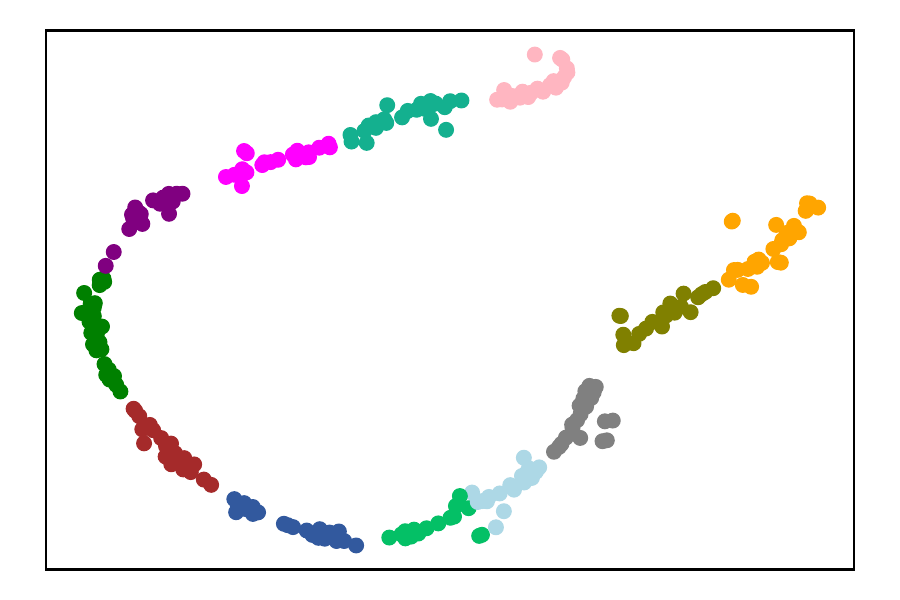}
		\caption{Visualization of \name.}
		\label{subfig:s0}
	\end{subfigure}
	\centering
	\begin{subfigure}{0.49\linewidth}
		\centering
		\includegraphics[width=\linewidth]{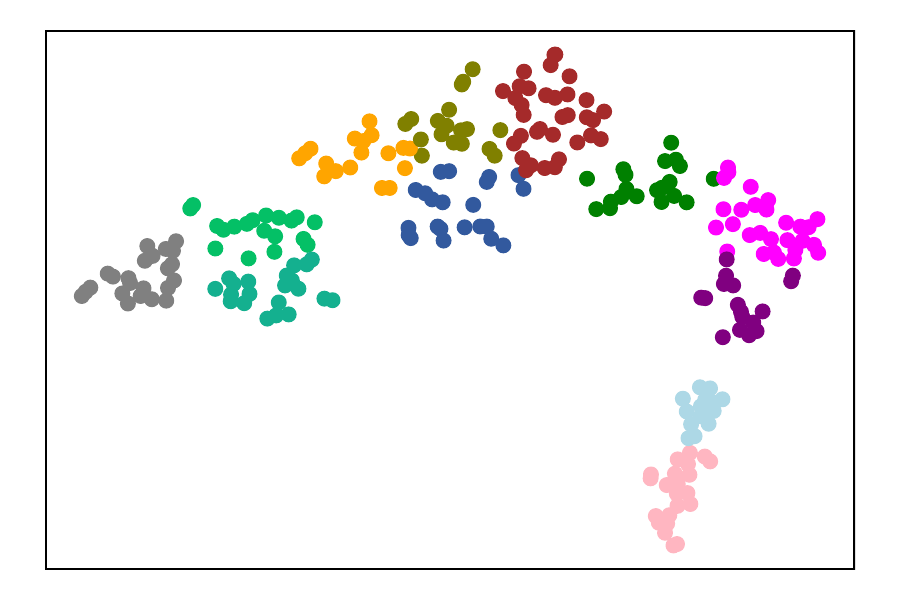}
		\caption{Visualization of \textit{w/o AM}.}
		\label{subfig:s1}
	\end{subfigure}
    \caption{t-SNE of $\mathbf{O}_{L,1}$ of \name and \textit{w/o AM} on GBA. Each color denotes a region-type cluster in the embedding space.}
	\label{fig:case}
\end{figure}

\subsection{Parameter Sensitivity}

We analyze the impact of four key hyperparameters in the GBA dataset. The results are illustrated in Figure~\ref{fig:param}.

\paragraph{Size of base weights $M$.} The adaptive mixing MLP generates region-specific matrices over a parameter pool with $M$ sets base weights. Smaller $M$ strengthens parameter sharing but limits adaptivity, while larger $M$ enriches the basis components for composition. However, overly large parameter pool can increase optimization difficulty and impair the model’s generalization ability.

\paragraph{Length of temporal window $p$.} The temporal aggregation mixer partitions the sequence into non-overlapping windows of length $p$ and performs within-window aggregation. Figure~\ref{fig:param} shows a clear degradation as $p$ increases. This behavior is consistent with the design motivation that local temporal proximity exhibits strong similarity, which can be effectively captured when $p$ is small. A larger $p$ may over-smooth short-term variations and discard high-frequency signals that are crucial for short-horizon prediction.

\paragraph{Number of adaptive region mixers $K$.} This determines the number of region scales in the spatial cascade mixer and controls the depth of spatial hierarchy. Moderate $K$ supports multi-scale regional semantics. When $K$ is small, the spatial hierarchy becomes insufficient, limiting ability to capture multi-scale regional semantics simultaneously. Excessively large $K$ introduces additional coarser aggregation levels that may inject extra noise.

\paragraph{Number of ST mixing blocks $L$.} This controls the depth of hierarchical spatiotemporal mixing and the resulting feature pyramids. Increasing $L$ expands the temporal receptive field, enabling higher layers to capture broader trends and to refine lower-level representations through top-down propagation.

\subsection{Visualization}
To evaluate \name's ability to capture regional semantics, we leverage t-SNE to visualize the spatial representations $\mathbf{O}_{L,1}$ produced by the \name and its \textit{w/o AM} variant. As illustrated in Figure~\ref{fig:case}, the representations generated by \name are more compact with well-defined boundaries between clusters. In contrast, the \textit{w/o AM} variant yields representations with substantial overlap between clusters. This comparison highlights \name's greater representation ability, which ensures nodes within the same region type share similar spatiotemporal patterns, while nodes across different types maintain distinct characteristics.

\section{Conclusion}
We present \name, a simple yet effective MLP-Mixer model, which leverages hierarchical modeling to tackle large-scale traffic forecasting. \name surpasses all current baselines and introduces enhancements to the MLP-Mixer architecture, making it competitive with more complex transformer-based models. Comprehensive experiments on extensive traffic datasets demonstrate its scalability and effectiveness. Future work will extend its application to multivariate time series prediction and anomaly detection.

\appendix



\section*{Acknowledgments}
This work is supported by the National Natural Science Foundation of China (No. 725B2004, 72242101, 72222022, 72171013), State Key Laboratory of Complex \& Critical Software Environment (SKLCCSE-2025ZX-17), and the Special Fund for Health Development Research of Beijing (2024-2G-30121).

\section*{Contribution Statement}
$^{\dagger}$ Yongyao Wang and Xie Yu contributed equally to this work.


\bibliographystyle{named}
\bibliography{ijcai26}

\end{document}